\documentclass[sigconf]{acmart}
\usepackage{booktabs} 
\usepackage{amsmath,url,graphicx,amssymb}
\usepackage{amsfonts}
\usepackage{ctable}
\usepackage{multirow}
\usepackage{algorithm}
\usepackage{algpseudocode}
\usepackage{pifont}
\usepackage{color}
\usepackage{subfigure}
\usepackage{enumitem}

\newcommand{\mylistbegin}{
  \begin{list}{$\bullet$}
   {
     \setlength{\itemsep}{-2pt}
     \setlength{\leftmargin}{1em}
     \setlength{\labelwidth}{1em}
     \setlength{\labelsep}{0.5em} } }
\newcommand{\mylistend}{
   \end{list}  }

\newcommand{\eg}{\textit{e.g.}}

\newcommand{\ie}{\textit{i.e.}}
\newcommand{\etc}{\textit{etc}}

\newcommand{\wrt}{\textit{w.r.t.~}}
\newcommand{\header}[1]{{\vspace{+1mm}\flushleft \textbf{#1}}}

\copyrightyear{2019}
\acmYear{2019} 
\setcopyright{iw3c2w3}
\acmConference[WWW '19]{Proceedings of the 2019 World Wide Web Conference}{May 13--17, 2019}{San Francisco, CA, USA}
\acmBooktitle{Proceedings of the 2019 World Wide Web Conference (WWW'19), May 13--17, 2019, San Francisco, CA, USA}
\acmPrice{}
\acmDOI{10.1145/3308558.3313456}
\acmISBN{978-1-4503-6674-8/19/05}

\begin{document}
\title{Place Deduplication with Embeddings}
\author{Carl Yang\footnotemark[1]{\small \footnotemark[2]}, Do Huy Hoang{\small \footnotemark[2]}, Tomas Mikolov{\small \footnotemark[2]}, Jiawei Han\footnotemark[1]}
       \affiliation{
       \institution{\footnotemark[1]University of Illinois, Urbana Champaign, 201 N Goodwin Ave, Urbana, IL 61801, USA}
       \institution{{\small \footnotemark[2]}Facebook Inc., 770 Broadway, New York, NY 10003, USA}
       \institution{\footnotemark[1]\{jiyang3, hanj\}@illinois.edu, {\small \footnotemark[2]}\{hoangmit, tmikolov\}@fb.com}
       }

\setlength{\floatsep}{4pt plus 4pt minus 1pt}
\setlength{\textfloatsep}{4pt plus 2pt minus 2pt}
\setlength{\intextsep}{4pt plus 2pt minus 2pt}
\setlength{\dbltextfloatsep}{3pt plus 2pt minus 1pt}
\setlength{\dblfloatsep}{3pt plus 2pt minus 1pt}
\setlength{\abovecaptionskip}{3pt}
\setlength{\belowcaptionskip}{2pt}
\setlength{\abovedisplayskip}{2pt plus 1pt minus 1pt}
\setlength{\belowdisplayskip}{2pt plus 1pt minus 1pt}
\renewcommand{\shortauthors}{Carl Yang, et al}
\settopmatter{printacmref=false, printfolios=false}

%!TEX root = dedup.tex
\begin{abstract}
Thanks to the advancing mobile location services, people nowadays can post about places to share visiting experience on-the-go. A large place graph not only helps users explore interesting destinations, but also provides opportunities for understanding and modeling the real world. To improve coverage and flexibility of the place graph, many platforms import places data from multiple sources, which unfortunately leads to the emergence of numerous duplicated places that severely hinder subsequent location-related services. 
In this work, we take the anonymous place graph from Facebook as an example to systematically study the problem of place deduplication: We carefully formulate the problem, study its connections to various related tasks that lead to several promising basic models, and arrive at a systematic two-step data-driven pipeline based on place embedding with multiple novel techniques that works significantly better than the state-of-the-art.
\end{abstract}
\keywords{place deduplication; feature generation; metric learning}
 
\maketitle
%!TEX root = dedup.tex
{\small \textbf{ACM Reference Format:}\\
Carl Yang, Do Huy Hoang, Tomas Mikolov, Jiawei Han. 2019. Place Deduplication with Embeddings. In \textit{Proceedings of the 2019 World Wide Web Conference (WWW'19), May 13-17, 2019, San Francisco, CA, USA.} ACM, New York, NY, USA, 7 pages. https://doi.org/10.1145/3308558.3313456}
\section{Introduction}
\label{sec:intro}
With the prevalence of mobile devices and online social platforms, maps and location related apps have become an integral part of people's everyday life. It has encouraged the development of various location-oriented infrastructures and services \cite{lian2014geomf, li2015rank, ye2011exploiting, liu2014exploiting, ference2013location, sarwat2014lars, yin2015joint}. 
For most of them, one key task is the maintenance of a high-quality map database (\textit{a.k.a.}~\textit{place graph}), which consists of various real-world places and their attributes. 
%To our knowledge, typical social network platforms nowadays maintain their place graphs of millions of place pages, which continue to grow with thousands of newly created ones every day. 
On one hand, such a database can support various map-based activities like \textit{search}, \textit{like}, \textit{check-in} and \textit{post}. Improving the efficiency and quality of such activities is beneficial to various stakeholders including business owners, advertisement agencies, and common users. On the other hand, the frequent location-based activity data generated by users provide an invaluable opportunity to improve the representation and understanding of the real world, and may further shed light on the capturing and modeling of various real-world activities.

However, since place pages are usually created from multiple sources whose qualities are hard to verify upon creation, the place graphs are often subject to large amounts of duplications, which hurts the quality of subsequent location-based services. For example, when a user touring in \textsf{New York City} wants to make a post at \textsf{Times Square}, if the system returns a list of places like \textsf{``New York Time Square''} \textsf{``Manhattan, Time Square''} \textsf{``TimeSquare, NYC''}, the user will get confused about which one to post on. Worse still, if the user also wants to explore restaurants or shopping malls around \textsf{Times Square}, then she/he may need to manually combine multiple lists of stores to get a full picture of all available choices.

Such user experiences are mainly due to the problem of duplicated places. 
To ease users' exploration of places and sharing of experiences, in this work, we take the \textit{massive} anonymous place graph from Facebook as an example to systematically study the problem of \textit{place deduplication}, and comprehensively evaluate the performance of different methods through extensive experiments on \textit{duplication candidate fetch} and \textit{pair-wise duplication prediction}. %Specifically, we propose a novel large-scale\textit{place embedding} pipeline that efficiently integrates \textit{ad-hoc place attributes} and \textit{noisy training signals} with deep neural networks. Our data-driven pipeline coherently combines the two steps of (1) \textit{unsupervised feature generation} and (2) \textit{supervised metric learning} to produce high-quality place embedding and efficiently support various downstream tasks.

Specifically, we start with \textit{large-scale} unsupervised feature generation to encode various ad-hoc place attributes through \textit{embedding learning} %\cite{mikolov2013efficient, mikolov2017advances} 
and \textit{network-based embedding smoothing}. %\cite{yang2017bridging, zhang2018camel}. 
Upon that, we develop an effective supervised metric learning framework to find the most useful features indicating place duplications. To fully leverage the noisy labeled data, we novelly pack the model with a series of techniques including \textit{batch-wise hard sampling}, \textit{source-oriented attentive training}, and \textit{soft clustering-based label denoising}. Extensive experiments on real-world place data show that our pipeline can outperform state-of-the-art industry-level baselines by significant margins, while each of our novel model components effectively pushes the limit of the overall pipeline.

Note that, although we focus on the example of Facebook data, our place deduplication pipeline is \textit{general} and ready to be applied to any online platform with place data or other deduplication tasks. Full implementation of our place embedding pipeline based on PyTorch will be released after the acceptance of this work.

%The main contributions and the organizations of this work are summarized as follows. 
%\begin{itemize}[leftmargin=*]
%\item In Section \ref{sec:problem}, we draw attention to the problem of place deduplication and propose to solve it through embedding, which fundamentally improves the quality of place graphs and promotes various location-related services.
%\item In Section \ref{sec:related}, we provide comprehensive studies on related research problems to generate intuitions and baselines for our place embedding pipeline. 
%\item In Section \ref{sec:data}, we describe the unique data of place deduplication, which lays the motivations of our data-driven embedding pipeline.
%\item In Section \ref{sec:pipeline}, we present our embedding pipeline that coherently combines large-scale unsupervised feature generation and novel supervised metric learning, together with extensive experiments on real-world place data to demonstrate its effectiveness.
%\item A quick summary is provided in Section \ref{sec:con}.
%\end{itemize}

%!TEX root = dedup.tex
\section{Problem Formulation}
\label{sec:problem}
\begin{figure}[h!]
\vspace{5pt}  
    \includegraphics[width=0.98\linewidth]{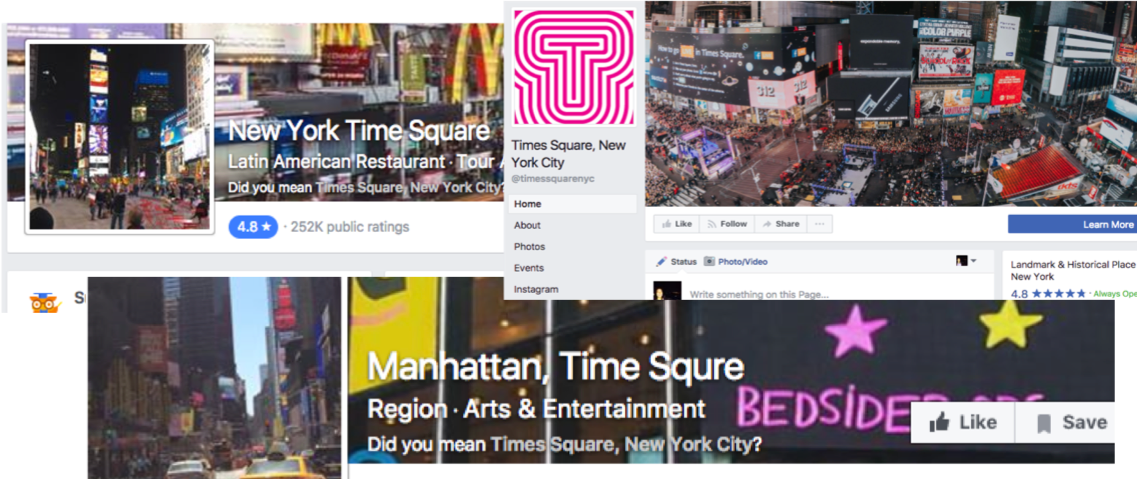}
%\vspace{5pt}    
    \caption{Example input with ad-hoc page features.}
    \label{fig:input}
\end{figure}

Nowadays, many online platforms maintain a \textit{place graph} to map the real world. 
The building blocks of such place graphs, however, instead of actual places, are usually \textit{online place pages} that capture certain attributes of actual places, as illustrated in Figure \ref{fig:input}.
Since such online places can be created by users, imported from third-party companies or scraped from the Web, many of them are \textit{duplicated}, as they in fact describe the same real-world places. 
In this work, to improve the quality of place graphs, we aim at the following novel but emergent problem.
\begin{definition}{Place Deduplication}.
Given a large set of online place pages, find those describing the same real-world places.
\end{definition}

\header{Input.} 
Given a set of online places $\mathcal{P}$, we represent the attributes of each place $p_i \in \mathcal{P}$ as $\mathbf{a}_i \in \mathcal{A}$. Such attributes can be ad-hoc, for example, place names, addresses, phone numbers, visitor histories, images, relevant posts and so on, which are essentially combinations of numerical, categorical, textual and even visual features. Moreover, due to the reality, many attributes can be missing and inaccurate.

\header{Output.}
For each place $p_i \in \mathcal{P}$, we aim to compute a unique fixed-sized low-dimensional vector (\textit{a.k.a.}~embedding) $\mathbf{u}_i \in \mathcal{U}$, to encode its ad-hoc attributes $\mathbf{a}_i \in \mathcal{A}$. %Thus, it is desirable that $\mathcal{U}$ captures the essentially useful information in $\mathcal{A}$ that distinguish duplicated and non-duplicated places, so as online places corresponding to the same actual places will lie close in the embedding space. %, while those describing different actual places will lie far apart.
%as illustrated in Figure \ref{fig:output}.
After getting such embeddings, various related tasks can be efficiently addressed through off-the-shelf machine learning algorithms. 
For example, duplication candidate fetch (find places that may be potentially duplicated) can be done through fast $k$-NN search \cite{garcia2008fast, JDH17}, pair-wise duplication prediction (decide if two places are duplications) simply involves the computation of pair-wise Euclidean distances, and meta-page discovery (find common actual places among all online places) can be achieved using clustering techniques such as $k$-means \cite{han2011data}.

%\begin{figure}[h!]
%    \includegraphics[width=0.9\linewidth]{figures/output.png}
%    \caption{Example output of page embeddings.}
%    \label{fig:output}
%\end{figure}

\header{Labels.}
To learn the proper place embedding $\mathcal{U}$ that facilitates place deduplication, besides place attributes $\mathcal{A}$, we also aim to leverage some labeled data $\mathcal{L}$, which are mainly from curation, crowdsourcing, and feedbacks. The labels are usually in the form of place pairs like $l=\{(a,b), y_l\}$, where $a$ and $b$ are two probe places. $y_{l}=1$ denotes a positive pair, meaning $a$ and $b$ are labeled as duplications, and $y_{l}=0$ denotes a negative pair. Labels from different sources can have varying qualities and biases.

\header{Present work.}
Taking the ad-hoc raw place attributes $\mathcal{A}$, we firstly transform them into fixed-sized numerical vectors $\mathcal{X}$ with unsupervised learning techniques, which captures the key information in $\mathcal{A}$ and can be easily processed by subsequent machine learning algorithms. Guided by labeled place pairs in $\mathcal{L}$, we then explore the feature representation $\mathcal{X}$ through supervised learning techniques and learn the final place embeddings $\mathcal{F}$ as vectors in a metric space where labeled duplicated places are close and non-duplicated places lie far apart. 
While we focus on place deduplication, the concepts and methods developed in this work are not bounded to the particular problem, but can be generally useful for deduplicating various objects like texts and users with similar types of attributes.

%\header{Impacts.}
%Motivated by recent advances in neural embedding \cite{chen2017beyond, schroff2015facenet, mikolov2013efficient, mikolov2013distributed}, we believe a proper place embedding framework can be useful in improving the quality of our place graph, so as to ease place exploration and experience sharing for users. 
%Particularly, through the precise representation of places with embedding, services like search relevance \cite{metzler2009improving} and vectorized suggestions \cite{kalantidis2013getting} can be naturally promoted, which potentially benefits different real-world stakeholders like business operators, advertisement agencies and common customers themselves. More broadly, learning the place representations allows us to efficiently describe and understand the physical world, while the embedding vectors act as bridges among different objects like places, people, events and so on. In this sense, place embedding is an indispensable part of the big vision of embedding and modeling everything in the whole world.
%!TEX root = dedup.tex
\section{Related Work and Motivations}
\label{sec:related}
We find the place deduplication problem closely related to several key tasks that heavily rely on advanced machine learning techniques, especially in the domains of computer vision (CV) and natural language processing (NLP).

\subsection{CV: Person Re-Identification}
One of the key tasks in CV is person re-identification \cite{chen2017beyond, ding2015deep, chen2016deep, cheng2016person, su2016deep, wang2016joint}, also referred to as face recognition \cite{schroff2015facenet, sun2015deeply, taigman2014deepface}. Given a large number of human images, the task is to find images of the same person. In concept, this task is quite close to ours as defined in Section \ref{sec:problem}. 

In CV, mainstream algorithms often tackle re-identification by projecting the images into a latent embedding space, while tuning the embedding space based on labeled data with a \textit{triplet loss} to enforce the correct order for each probe image. Their conceptual optimization objective can be expressed as follows.
\begin{align}
\mathcal{J}_{tr} = \sum_{t=(a,p,n)} &\text{max}(0, d(\mathbf{u}_a, \mathbf{u}_p)-d(\mathbf{u}_a, \mathbf{u}_n)+\alpha),
\label{eq:trip}
\end{align}
where $\mathbf{u}=\mathbf{f}(\mathbf{x})$ and $d(\mathbf{u}_i, \mathbf{u}_j) = d_e(\mathbf{u}_i, \mathbf{u}_j) = ||\mathbf{u}_i-\mathbf{v}_j||_2^2$. $\mathbf{x}$ is the original image features (\eg, $96\times96$ pixels), $\mathbf{f}(\cdot)$ denotes the embedding projection function (\eg, a deep CNN), $\alpha$ is the margin hyper-parameter, and $||\cdot||_2^2$ denotes the Frobenius norm. 
%To make the learning process more efficient, hard sampling is often introduced, to put more weights on samples with lager losses during training.

However, while images naturally come with rich visual features that can be well explored by the CNN models, the attributes of our places are much more ad-hoc and hard to capture. 
%It makes the direct adoption of person re-identification models into our scenario impossible and motivates us to find a systematic solution for place feature generation, which we will talk about in Section \ref{sec:unsupervised}.
Another limitation of their models is the requirement of training data to be accurate and in the form of \textit{triplets}, \ie, $t=(a, p, n)$, where $a$ is the probe image, $p$ is the positive image and $n$ is the negative image. In person re-identification, the labels are often generated from image collections taken of the same persons. Therefore, while the quality of input images may be different, the labels are often accurate, and the triplet requirement is easy to be satisfied.

In our scenario of place deduplication, labels are generated from different sources like curation and crowdsource based on samples of all possible place pairs, and are thus prone to varying qualities and biases. 
Moreover, instead of triplets, the training data naturally come as \textit{pairs}, in the form of $l=\{(a, b), y_{l}\}$, as defined in Section \ref{sec:problem}. 
Finally, place duplication labels are extremely sparse-- in our case, only about $10^{-5}$ percent of place pairs are labeled. %, and less than $1\%$ of the labeled pairs form natural triplets.
It requires non-trivial adaptions of the CV models to work for place deduplication, which we will talk about in Section \ref{sec:unsupervised} and \ref{sec:supervised}.

%\begin{table*}[t!]
%\small
%\centering
%\setlength\tabcolsep{2pt}
%\begin{tabular}{|c|c|c|c|}
%\hline
%\textbf{name} & \textbf{address} & \textbf{coordinate} & \textbf{category}\\ 
%\hline
%Metropolitan Museum of Art & 1000 5th Ave, New York, NY 10028 & 40.7887, 73.9638 &tourist site \\
%\hline
%The MET & New York & -, - & tourist site \\
%\hline
%Central Park & New York & 40.7728, 73.9645 & - \\
%\hline
%Mecy's & 151 W 34th St, New York, NY 10001 & 40.7512, 73.9880 & shopping \\
%\hline
%Mecy's & 422 Fulton St, New York & 40.6908, 73.9857 & shopping\\
%\hline
%\end{tabular}
%\caption{\label{tab:attribute} Examples of a few places in the anonymous Facebook place graph.\\
%Only the first two are duplicated places.}
%\end{table*}

\subsection{NLP: Entity Resolution}
%Entity resolution, as a crucial task of NLP to find different mentions to the same entities, is also highly related to our task.

If we only consider place names, the problem is close to synonym detection. However, as most algorithms leverage sentence-level contexts to learn word similarities \cite{ustalov2017watset, qu2017automatic, roller2014inclusive, weeds2014learning, qian2009semi, sun2010semi}, they are not directly applicable to place deduplication where such contexts are often not available.
Instead, we may consider direct matching of place names, which can be done using deep neural networks \cite{hu2014convolutional, wan2016deep, sun2018multi, guo2016deep, xiong2017end, pang2016text, chen2018mix}. However, these models are known to be more suitable for longer sentences with more complex structures, rather than the short simple place names, and they require large amounts of training data, which is extremely hard to acquire for place deduplication \cite{christen2015efficient, dal2016practical, qian2017active}.
Finally, by considering additional place attributes like location, ad-hoc models have been developed based on heuristic feature engineering \cite{dalvi2014deduplicating, chuang2015verification, kozhevnikov2016comparison}, but their flexibility is limited to take more different place attributes, such as address and category. Thus, it is desirable for us to develop our own systematic and flexible place deduplication pipeline beyond the existing works on entity resolution, as we will discuss in Section \ref{sec:unsupervised} and \ref{sec:supervised}.

Interestingly, another particular trend of entity resolution is to construct the word networks from search engine query logs and compute network clustering to detect synonym sets \cite{he2016automatic, ren2015synonym, chakrabarti2012framework}. Since it is non-trivial to put everything we have into a single network, their methods are not directly applicable. Nonetheless, we find the ideas of network construction and clustering intriguing, and are able to leverage both of them in Section \ref{sec:unsupervised} and \ref{sec:supervised}, respectively.

\section{Data and Experimental Setups}
\label{sec:data}
In this section, we describe the massive dataset we use, upon which place deduplication is desired. It is a subset of the anonymous real-world place graph from Facebook including a total of 730M online place pages.
In this work, we take the Facebook place graph as an example to show the effectiveness and efficiency of the methods we develop, while they are generally applicable to any online platforms with place data or other deduplication tasks. 
\subsection{Data Preparation}
As we discussed in Section \ref{sec:problem}, places can have various ad-hoc attributes with varying qualities and incompleteness, which need to be leveraged through specifically designed methods, as we will show in Section \ref{sec:unsupervised}. 
%We will also briefly discuss how to capture other features, but leave them for future work.
%characteristics of features are deferred for motivating the methods

We process and generate training data of about 4.6M pairs of places from 19 different sources, including \textsf{curation}, \textsf{crowdsourcing}, \textsf{feedbacks} and so on. Labels from curation are generally of higher qualities, while most of the others like feedbacks generated through surveys and crowdsourcing produced by cheap labors may be rather \textit{noisy}. They cover a total of about 4M places. Among them, about 1.1M pages are labeled with duplications and 3M pages with non-duplications. The average number of duplications is 0.35, whereas the maximum number of duplications is 297. The average number of non-duplications is 1.98, and the maximum number of non-duplications is over 1K. Therefore, the training data are \textit{sparse} and \textit{biased}, and such sparsity and bias are different across the sources. In Section \ref{sec:supervised}, we will discuss how to leverage such training data with varying noise, sparsity, and bias.
%discussion of sources· are deferred for motivating the methods

\subsection{Evaluation Methods}
The testing data we use is a golden set of 47K pages completely separate from the training data, which is deliberately curated for place quality evaluation for production. It contains different place pages from the training data.
%Among them, 11K are labeled with duplications, while 36K are labeled with non-duplications. The average duplication and non-duplication numbers are 0.24 and 0.89, respectively, which are significantly lower than the training data. 

%In this work, we focus on the comparison of different embedding frameworks for place deduplication and comprehensively evaluate their performance through multiple metrics on two key tasks for place deduplication: pair-wise duplication prediction and point-wise duplication candidate fetch. 

To compare different embedding models, we firstly compute the vector representations $\mathcal{U}$ of all places in the testing set. For the set of places labeled with both duplications and non-duplications (\ie, $a \in \Omega$), we compute the average pair-wise accuracy as follows:
\begin{align}
\text{ACC}=\frac{1}{|\Omega|}\sum_{a\in\Omega}\frac{1}{|\Phi_a||\Psi_a|}\sum_{p\in\Phi_a, n\in\Psi_a}\mathbb{I}(||\mathbf{u}_a-\mathbf{u}_p||_2^2 < ||\mathbf{u}_a-\mathbf{u}_n||_2^2),\nonumber
\end{align}
where $\Phi_a$ and $\Psi_a$ are the sets of labeled duplications and non-duplications of place $a$, respectively. %and $\mathbb{I}(\cdot)$ is the indicator function that takes 1 when the argument inside is true, and 0 otherwise. 
This metric measures the utility of the embedding for the task of pair-wise duplication prediction.

Besides ACC, we also compute the nearest neighbors of each place through fast $k$-NN \cite{JDH17}, and compute the precision and recall at $K$ based on the set of places labeled with duplications (\ie, $a \in \Theta$), to measure the utility of the embedding for the task of point-wise duplication candidate fetch.% as follows:
%\begin{align}
%\text{PRE@K}=\frac{1}{|\Theta|}\sum_{a\in\Theta}\frac{\sum_{k=1}^K \mathbb{I}(b_{a,k} \in \Phi_a)}{K},\\
%\text{REC@K}=\frac{1}{|\Theta|}\sum_{a\in\Theta}\frac{\sum_{k=1}^K \mathbb{I}(b_{a,k} \in \Phi_a)}{|\Phi_a|},
%\end{align}
%where $b_{a,k}$ is the $k$-th nearest neighbor of $a$ in the embedding space. We vary $K$ from 1 to 100 to provide a complete evaluation. These metrics measure the utility of the embedding for the task of point-wise duplication candidate fetch.

%\subsection{State-of-the-art Baselines}
%A major advantage of place embedding is it facilitates fast nearest neighbor search for duplication candidate fetch. On the contrary, the state-of-the-art industry-level solution for place deduplication is to fit a supervised learning model (\eg, random forest or gradient boosting decision tree) on labeled pairs of duplicated or different places. Such models may be effective in pair-wise duplication prediction on particular place pairs, but are too slow to be useful for point-wise duplication candidate fetch in large spaces. Therefore, we only compute and present the ACC scores for them in Section \ref{sec:supervised} to show the superiority of our proposed embedding-based methods.

%!TEX root = dedup.tex
\section{Unsupervised Feature Generation}
\label{sec:unsupervised}
The key place attributes we aim to leverage in this work are name, address, coordinate and category, while other attributes can also be easily incorporated in the future. Among them, name and address are textual, coordinate is numerical, and category is categorical, which need to be handled differently. Moreover, many of them are noisy and incomplete, requiring the model to be robust and flexible. 

\subsection{Encoding Name}
We start with place name, because it is often the primary and most indicative feature towards deduplication%-- without other features, by looking at the names of two places, one can often tell if they are duplications 
(\eg, \textsf{``Metropolitan Museum of Art''} and \textsf{``The MET''}). Motivated by the success of word embedding in NLP, we propose to capture the place name semantics through embedding. The idea is to leverage the distributional information about words and infer word semantic similarities based on the context, so as to recognize common misspellings (\eg, \textsf{``capitol''} as \textsf{``capital''}, \textsf{``corner''} as \textsf{``conner''}), acronyms (\eg, \textsf{``street''} as \textsf{``st''}), synonyms (\eg, \textsf{``plaza''} and \textsf{``square''}) and \etc. %To this end, we propose and evaluate two place name embedding methods.

\header{Method 1: \textit{Skip-gram} + place name corpus.}
We firstly train a simple word embedding model on all place names we have, which include 730M short texts. For preprocessing, we prune all place names into the shortest name variants by removing all location prefixes and suffixes in the names (\eg, remove \textsf{``New York''} from \textsf{``Time Square New York''}); we also normalize all place names by replacing special characters with spaces (\eg, replace ``\$'' or emojis with a single space) and change all letters to lower cases. %After these steps, the average number of words in place names is 3.11.

Since place names are often quite short (on average only 3.11 words after preprocessing), we directly apply the \textit{Skip-gram} model \cite{mikolov2013distributed} by sampling word-context pairs from place names. %Other parameters are set as default in the open source library\footnote{https://www.tensorflow.org/versions/r0.12/tutorials/word2vec/}. 
Place embedding is then computed as the average of word embedding.
%After getting the word embeddings, we compute the place name embedding by averaging the embeddings of all words in the names.

This method, while efficiently providing numerical place name representations that can be processed by subsequent machine learning algorithms, has quite a few drawbacks as listed below.
\begin{itemize}[leftmargin=*]
\item Limited distributional information: The training corpus, although is quite large, provides rather limited statistics around words and their contexts, since the corpus is chunked into short texts.% and there is no good order to aggregate them. %This is especially concerning when the whole place names are misspelled (\eg, ``\textsf{stabucks}'') or abbreviated (\eg, ``\textsf{The MET}''), where there is hardly any useful context for the particular words.
\item Biased samples: Unlike sampling word-context pairs from fixed-sized sliding windows, when sampling from the place names, it is hard to avoid the bias towards either shorter or longer names. %Moreover, the frequency and position distributions of words in place names are different from natural language, which may lead to bias towards certain words (\eg, the word \textsf{``park''} appears much more often in place names than in natural language, and is often at the end of the place name).
\item Ignorance of word internal structure: Standard word vectors ignore word internal structure that contains rich information, which might be especially useful for rare or misspelled words.% (\eg, \textsf{``money''} as \textsf{``m0ney''}).
\end{itemize}

\header{Method 2: \textit{Fast-text} + Facebook post corpus.}
To ameliorate the above limitations, we adopt the advanced \textit{Fast-text} method \cite{mikolov2017advances}, with strategies including position-dependent weighting, phrase generation and subword enrichment \cite{bojanowski2017enriching, joulin2017bag}. Moreover, to capture more casual languages used by social network users who frequently create the place pages, instead of training the \textit{Fast-text} model on standard NLP corpus like Wikipedia\footnote{https://dumps.wikimedia.org/enwiki/latest/}, % and news\footnote{https://www.ldc.upenn.edu/language-resources/data}, 
we use anonymized word embeddings derived from the public posts on Facebook in 10 years. The corpus has around 1.9T words in total, which is about 3 times bigger than the massive Common Crawl data\footnote{https://commoncrawl.org/2017/06} used to train the state-of-the-art public word embedding \cite{mikolov2017advances}.
%Specifically, we use the CBOW with negative sampling in \cite{mikolov2013efficient} as the basic model to capture the distributional information about words. We successfully scale up the model and generate a 300-dim embedding over 8M distinct words in 4 days. The corpus we use provide plentiful distributional information even for the relatively rare words. Training with the Facebook post corpus also helps reduce much of the sampling bias, which is further mended through frequent word down-sampling \cite{mikolov2013distributed} and position-dependent weighting \cite{mnih2013learning}. 
%To better utilize the distributional information, we enrich the model by generating phrases \cite{mikolov2017advances} and considering subwords \cite{bojanowski2017enriching}, which further improves the embedding of rare words, misspellings, and abbreviations. %The embeddings of place names are the average of embeddings of the words they contain.

\subsection{Incorporating Address}
In our place graph, about two-thirds of the places have address information. It is especially useful in differentiating branches of the same stores (\eg, the different \textsf{Starbucks} in a city). Compared with names, addresses are more often incomplete, but less noisy, because most of them are validated and filtered before recorded. Therefore, we observe much fewer misspellings and abbreviations. Moreover, addresses are often longer than names, which naturally provides richer distributional information and allows less biased samples. Finally, in addresses, we do not really want semantically similar words to be close-- for example, different words like \textsf{``street''}, \textsf{``avenue''} and \textsf{``boulevard''}  in the address should indicate different places although their semantic meanings are similar. 
We use the same two approaches for place names to embed place addresses.
%Due to these reasons, we propose to not use the \textit{Fast-text} embedding computed on Facebook post corpus for address representation. Instead, we should only train a simple \textit{Skip-gram} model on all available addresses similarly as described in Method 1 above, and obtain the 50-dim address embedding for places that have addresses. A 50-dim zero vector is used for places that do not have addresses.

\subsection{Leveraging Coordinate and Category}
Place coordinates and categories provide additional information about duplications. Coordinates are just 2-dim numerical vectors, whereas categories can be converted to 0-1 dummy variables. %Since there are so many categories and some are meaningless, 
In this work, we focus on 19 common ones such as \textsf{Shopping} and \textsf{Restaurant}, while the algorithm can generalize to any other appropriate subsets of categories.

A simple way to leverage place coordinate and category is to concatenate the 2-dim numerical coordinate and 19-dim 0-1 dummy vector of category to the place embedding. However, in practice, we find such concatenation not helpful and even lead to worse performance as can be seen later, probably because such variables are not compatible with the word embeddings. 

Motivated by a recent work on place recommendation \cite{yang2017bridging}, we refine the place embedding through unsupervised embedding smoothing on a place network. The idea is to require the embeddings of places that have similar coordinates or same categories to be close. 

Specifically, we construct a place network $\mathcal{N}=\{\mathcal{P}, \mathcal{E}\}$, where $\mathcal{P}$ is the set of all places. We then construct two types of edges $\mathcal{E}=\mathcal{E}_1\cup\mathcal{E}_2$ based on coordinates and categories, respectively. Following the grid-based binning approach in \cite{yang2018did}, we group places into squared bins based on coordinates, and add a coordinate edge $e_i \in \mathcal{E}_1$ between places in the same bins; a category edge $e_j \in \mathcal{E}_2$ is added between places belonging to the same categories. 

Following \cite{yang2017bridging}, we derive the loss that enforces smoothness among places that are close on the place network as
\begin{align}
\mathcal{J}_{sm}=&-\sum_{(p_i, p_c)}\log p(p_c|p_i)\nonumber\\
=&-\sum_{(p_i, p_c)} \log [\phi^T_c \mathbf{g}(\mathbf{a}_i)-\log \sum_{p'_c\in \mathcal{P}}\exp(\phi^T_{c'}\mathbf{g}(\mathbf{a}_i))]\nonumber\\
=&-\mathbb{E}_{(p_i, p_c, \gamma)}\log\sigma(\gamma\phi^T_c\mathbf{g}(\mathbf{a}_i)).
\label{eq:sm}
\end{align}

In the first line, we firstly formulate the standard \textit{Skip-gram} objective adapted to predict the correct graph context of place based on coordinates and categories, which is then decomposed into the second line, where $\phi_c$ is the learnable context embedding of place $p_c$, $\mathbf{g}(\cdot)$ is a learnable smoothing function (\eg, a single layer perceptron with the same input and output sizes) that maps the original place embedding $\mathbf{a}_i$ to the smoothed embedding space. This objective is hard to estimate due to the summation over all contexts in $\mathcal{P}$ in the second term, so we write it into an equivalent expectation over the distribution of $p(p_i, p_c, \gamma)$ in the third line, where $\sigma(x)=1/(1+e^{-x})$, and approximate it by applying the popular negative sampling approach \cite{mikolov2013distributed}.

\subsection{Experimental Evaluations}
We compare the following combinations to comprehensively show the overall effectiveness of our framework as well as how each of the model components helps in improving the embedding quality.
\begin{itemize}[leftmargin=*]
\item \textbf{NS:} 50-dim place \textit{Name} embedding produced by the \textit{Skip-gram} model trained on the place name corpus.
\item \textbf{NF:} 300-dim place \textit{Name} embedding produced by the \textit{Fast-text} model trained on the Facebook post corpus.
\item \textbf{NF+AS:} \textit{NF} concatenated with the 50-dim place \textit{Address} embedding produced by the \textit{Skip-gram} model trained on the place address corpus.
\item \textbf{NF+AF:} \textit{NF} concatenated with the 300-dim place \textit{Address} embedding produced by the \textit{Fast-text} model trained on the Facebook post corpus.
\item \textbf{NF+AS+CC:} \textit{NF+AS} \textit{Concatenated} with the 21-dim place \textit{Coordinate and category} vectors.
\item \textbf{NF+AS+CS:} \textit{NF+AS} \textit{Smoothed} on the place network constructed \wrt~place \textit{Coordinate and category}.
\end{itemize}

\begin{figure}[h!]
\vspace{-10pt}
\centering
\subfigure[PRE@K]{
\includegraphics[width=0.23\textwidth]{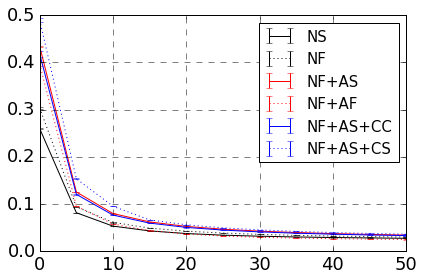}}
\hspace{-8pt}
\subfigure[REC@K]{
\includegraphics[width=0.23\textwidth]{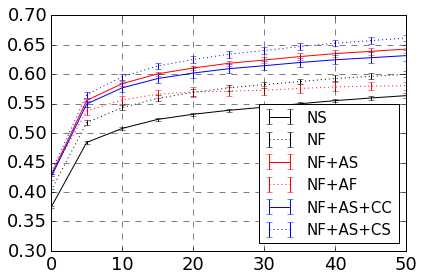}}
\vspace{-5pt}
\caption{\textbf{Performance of feature generation methods.}}
\vspace{-5pt}
\label{fig:unsup}
\end{figure}

\begin{table}[t!]
\small
\centering
\begin{tabular}{|c|c|c|c|}
\hline
\textbf{Method} & \textbf{avg.~PRE} & \textbf{avg.~REC} & \textbf{ACC}\\ 
\hline
\textbf{NS} & $.03794 \pm .00010$ & $.5334 \pm .0034$ & $.6212 \pm .0325$ \\
\hline
\textbf{NF} & $.04553 \pm. 00018$ &  $.5708 \pm .0052$ & $.6718 \pm .0107$ \\
\hline
\textbf{NF+AS} & $.05216 \pm .00030$ &  $.6162 \pm .0067$ & $.6917 \pm .0181$ \\
\hline
\textbf{NF+AF} & $.05008 \pm .00018$ &  $.5866 \pm .0061$ & $.6424 \pm .0324$\\
\hline
\textbf{NF+AS+CC} & $ .05026\pm.00034 $ &  $.6036 \pm. 0083$ & $.6716 \pm .0297$ \\
\hline
\textbf{NF+AS+CS} & $ .05508\pm.00024 $ &  $ .6476\pm .0058$ & $ .6989\pm .0135$ \\
\hline
\end{tabular}
\caption{\label{tab:unsup} \textbf{Performance of feature generation methods.}}
\vspace{-5pt}
\end{table}

Figure \ref{fig:unsup} and Table \ref{tab:unsup} show the performance of different feature generation methods. As can be clearly observed: 1) When we only consider the place name, \textit{NF} performs significantly better than \textit{NS} regarding all metrics, which supports our idea of training the advanced \textit{Fast-text} model on the massive Facebook post corpus;
%to effectively address the drawbacks of training the simple \textit{Skip-gram} model on the place name corpus; 
2) Incorporating place address is generally helpful, but the simple \textit{AS} model we advocate for produces more significant improvements than the heavy \textit{AF} model, which tells the importance of choosing the proper model rather than the complex one; 
3) To leverage place coordinate and category, our proposed network-based embedding smoothing method (\textit{CS}) is much more effective than the direct vector concatenation (\textit{CC}).

%Note that, in these experiments, we mainly look at the PRE and REC scores to understand the quality of place features generated in the unsupervised fashion, while in the next section, we will show that incorporating supervision from labeled training data can largely improve the ACC scores.
The \textit{avg.}~PRE and \textit{avg.}~REC are taken over the 100 measures of PRE@$K$ and REC@$K$ when $K$ varies from 1 to 100, to directly compare and reveal the relative effectiveness of different methods. The absolute values of \textit{avg.}~PRE are pretty low because the numbers of labeled duplications are low in our testing data.
Based on such results, we will use \textit{NF+AS+CS} as our feature generation model and use its output $\mathcal{X}$ as the input of subsequent supervised metric learning models.
Note that, other than places with the particular attributes as we focus on in this work, the methods developed here can be combined in different ways to capture textual, categorical and numerical attributes of various real-world objects.

%!TEX root = dedup.tex
\section{Supervised Metric Learning}
\label{sec:supervised}
Besides place attributes, we also aim to leverage labeled place pairs through supervised metric learning. The idea is to learn a non-linear projection function $\mathcal{F}$, which transforms the place features $\mathcal{X}$ into a metric space, where duplicated places are close and non-duplicated places are far apart. Therefore, the model should be able to explore and stress the important features that indicate duplications. %Moreover, as we discuss in Section \ref{sec:related} and \ref{sec:data}, the model should also efficiently deal with the \textit{noisy}, \textit{sparse} and \textit{biased} training data from \textit{multiple sources} in the form of \textit{pairs}.

\subsection{Basic Models}
We construct the basic metric learning models by applying simple MLP upon the place embedding $\mathcal{X}$. However, since direct leverage of standard triplet loss commonly used in person re-identification \cite{schroff2015facenet} and entity resolution \cite{qu2017automatic} in Eq.~\ref{eq:trip} requires the sampling of the third place based on our pair-wise place labels, which can introduce a lot of non-relevant and trivial training data, we propose to replace the triplet loss with a pair-wise contrastive loss as
\begin{align}
\mathcal{J}_{pr} = & \sum_{l=\{(a,b),y_l\}} [y_{l}d(\mathbf{u}_a, \mathbf{u}_b)
+(1-y_{l})\text{ max}(0, \alpha-d(\mathbf{u}_a, \mathbf{u}_b))],\nonumber
\label{eq:pair}
\end{align}
where $d=d_e$ is the same as in Eq.~\ref{eq:trip}. Besides the Euclidean distance function, we also tried various other distance functions including cosine similarity and learnable bilinear distance \cite{qu2017automatic}.

We comprehensively evaluated the performance of various basic models, and found that the \textit{PE} (Pair-wise loss with Euclidean distance function) constantly performs the best, indicating the effectiveness of replacing the standard triplet loss in traditional CV and NLP tasks. Based on such observation, we will focus on further improving the \textit{PE} model in the rest of this work.

\subsection{Hard Sampling}
Training the model with our massive noisy labeled place pairs can take quite significant time, most of which is wasted on trivial samples. To this end, we design the following novel criteria for dynamic batch-wise hard sampling towards our pair-wise loss.
%\begin{align}
%C_{t}= & \;\mathbb{I}\;[d(\mathbf{f}(\mathbf{x}_a), \mathbf{f}(\mathbf{x}_p))-d(\mathbf{f}(\mathbf{x}_a), \mathbf{f}(\mathbf{x}_n)) \nonumber\\
%& > \frac{\beta}{|B|}\sum_{t'\in B}d(\mathbf{f}(\mathbf{x}_a'), \mathbf{f}(\mathbf{x}_p'))-d(\mathbf{f}(\mathbf{x}_a'), \mathbf{f}(\mathbf{x}_n'))],
%\end{align}
\begin{align}
C_{l=(a,b)}&= \;\mathbb{I}\;[y_{l}d(\mathbf{u}_a, \mathbf{u}_b)-(1-y_{l})d(\mathbf{u}_a, \mathbf{u}_b) >  \nonumber\\
& \frac{\beta}{|B|} \sum_{l'\in B} y_{l'}d(\mathbf{u}_{a'}, \mathbf{u}_{b'})-(1-y_{l'})d(\mathbf{u}_{a'}, \mathbf{u}_{b'})],
\end{align}
where $B$ denotes the set of samples in a batch, $\beta$ is the slack hyper-parameter controlling the amount of hard samples, and $d(\cdot)$ can be implemented with either $d_f(\cdot)$ or $d_b(\cdot)$. A sample pair $l$ is selected to contribute to the loss for gradient computation when $C_l=1$. The idea is to only focus on the hard samples where the model fails to put the labeled duplications close enough or non-duplications far away, compared with the average distances. 

\subsection{Attentive Training}
Our labeled data are from different sources, which may have varying quality and bias. For example, data curated by a particular team at a certain time might be biased towards specific metrics. This often happens when people recognize the ignorance of particular situations and aim to improve them upon iterating to the next period of curation. Moreover, data labeled by the professional curation teams and rookie crowdsourcing workers can have quite different qualities. Therefore, it is intuitive to automatically assign different weights to the samples from different sources during training.

To deal with the varying quality and bias of training samples from multiple sources, we design a novel source-oriented attentive training technique based on the idea of self-attention \cite{denil2012learning, vaswani2017attention}, which has shown to be effective in various deep learning scenarios. 

%\begin{figure}[h!]
%    \includegraphics[width=0.9\linewidth]{figures/attentive.png}
%    \caption{Our novel attentive training framework.}
%    \label{fig:att}
%\end{figure}

For each place $p_i$, rather than learning a single embedding vector $\mathbf{f}(\mathbf{x}_i)$, we learn two embedding vectors $\mathbf{k}(\mathbf{x}_i)$ and $\mathbf{v}(\mathbf{x}_i)$, which we call the \textit{key embedding} and \textit{value embedding}, respectively. In practice, we can implement both $\mathbf{k}(\cdot)$ and $\mathbf{v}(\cdot)$ as FNNs and let them share a few layers. Besides, we also learn a global source embedding $\mathbf{q}(s_j)$ for each training source $s_j$, which can be implemented as an embedding look-up table. The size of $\mathbf{q}(\cdot)$ is twice as $\mathbf{k}(\cdot)$. Subsequently, we design the weight of the training sample $l=(a, b)$ from source $s$ as
\begin{align}
w_{ls} = \frac{\exp([\mathbf{k}(\mathbf{x}_a), \mathbf{k}(\mathbf{x}_b)]^T \mathbf{q}(s))}{\sum_{s'} \exp([\mathbf{k}(\mathbf{x}_a), \mathbf{k}(\mathbf{x}_b)]^T \mathbf{q}(s'))},
\end{align}
where $[\cdot]$ denotes vector concatenation. With $\mathcal{W}$ defined, we rewrite our loss function in Eq.~\ref{eq:pair} into
\begin{align}
\mathcal{J}_{pr}' = & \sum_s \sum_{l=\{(a,b),y_l\}\in \mathcal{T}_s} w_{ls} [y_{l}d(\mathbf{v}(\mathbf{x}_a), \mathbf{v}(\mathbf{x}_b))\nonumber\\
&+(1-y_l)\text{ max}(0, \alpha-d(\mathbf{v}(\mathbf{x}_a), \mathbf{v}(\mathbf{x}_b)))],
\label{eq:att}
\end{align}
where $\mathcal{T}_s$ denotes the set of training samples from source $s$.

Training with Eq.~\ref{eq:att} allows the model to learn a source-oriented self-attention weight for each training pair. To be specific, if similar samples keep appearing in a certain source, they are likely redundant and less useful, and the model will automatically reduce their contribution to the loss. However, if similar samples appear across different sources, the model will put more trust on their correctness by relatively increasing their attention weights. Finally, unseen samples will likely get higher attention. %To prevent the source embedding $Q$ from vanishing, we randomly initialize it to be a column-wise unit matrix $Q_0$ and apply an additional regularization loss as follows, where $\lambda$ is a weighting hyper-parameter
%\begin{align}
%\mathcal{J}_{reg} = \lambda||Q-Q_0||^2_2.
%\end{align}

\subsection{Label Denoising}
Place deduplication can be essentially regarded as a semi-supervised clustering problem, where pages of the same places should naturally form clusters in the embedding space. This is true mainly because the duplication relationship is \textit{transitive}. While such clusters are ideal for duplication detection, they can also help reduce the noise within training data, \ie, labels that conflict with the cluster structures are likely noisy. To leverage this insight, we design a novel soft clustering-based label denoising module based on the idea of self-training neural networks \cite{nigam2000analyzing, xie2016unsupervised, guo2017improved}. Particularly, we introduce a loss based on the KL-divergence
\begin{align}
\mathcal{J}_{dn} = \rho\;\text{KL} (\mathcal{C}||\mathcal{D})=\rho\sum_i\sum_k c_{ik} \log\frac{c_{ik}}{d_{ik}},
\end{align}
where $\rho$ is a hyper-parameter. $d_{ik}\in\mathcal{D}$ is the probability of assigning place $p_i$ to the $k$th cluster, under the assumption of Student's $t$-distribution with degree of freedom set to 1 \cite{maaten2008visualizing}, \ie, 
\begin{align}
d_{ik} = \frac{(1+||\mathbf{f}(\mathbf{x}_i)-\mathbf{u}_k||^2)^{-1}}{\sum_j(1+||\mathbf{f}(\mathbf{x}_i)-\mathbf{u}_j||^2)^{-1}}.
\end{align}
It is basically a kernel function that measures the similarity between the embedding of place $p_i$ and the cluster center $u_k$. $\mathcal{C}$ is an auxiliary target distribution defined as
\begin{align}
c_{ik} = \frac{d_{ik}^2/g_k}{\sum_{k'} d_{ik'}^2/g_{k'}},
\end{align}
where $g_k=\sum_i d_{ik}$ is the total number of places softly assigned to the $k$th cluster. Raising $\mathcal{D}$ to the second power and then dividing by the number of places per cluster allows the target distribution $\mathcal{C}$ to improve cluster purity and leverage the confident assignments to help reduce the influence of noisy labels.

\subsection{Experimental Evaluations}
We compare the following methods to demonstrate the effectiveness of our proposed supervised metric learning techniques.
\begin{itemize}[leftmargin=*] 
\item \textbf{NF+AS+CS:} The best place features we got through unsupervised feature generation in Section \ref{sec:unsupervised}, which is also the input of the following metric learning methods. 
\item \textbf{PE:} The best basic metric learning method adapted from CV and NLP with the \textit{Pair-wise contrastive loss} and \textit{Euclidean distance function}, as discussed in Section \ref{sec:supervised}.1.
\item \textbf{PEH:} \textit{PE} improved with batch-wise hard sampling.
\item \textbf{PEHA:} \textit{PEH} improved with our novel source-oriented attentive training technique.
\item \textbf{PEHAD:} \textit{PEHA} improved with our novel soft clustering-based label denoising technique.
\end{itemize}
We also compare our models with several state-of-the-art classification algorithms that can be applied for pair-wise duplication prediction, \ie, logistic regression (\textit{LR}), support vector machine (\textit{SVM}), random forest (\textit{RF}) and gradient boosting decision tree (\textit{GBDT}). These algorithms do not produce place embeddings and thus cannot be efficiently evaluated through fast $k$-NN in datasets with hundreds of millions of places. Therefore, we only compute their standard classification accuracy of duplication prediction on the testing data, as shown in Table \ref{tab:baseline}.

\begin{table}[t!]
\small
\centering
\begin{tabular}{|c|c|c|c|c|c|}
\hline
\textbf{Method} & LR & SVM & RF & GBDT & \textbf{PEHAD}\\ 
\hline
%\textbf{ACC} & $.7514\pm.0011$ & $.7699\pm.0023$ & $.7786 \pm.0018$ & $.7828\pm.0010$ \\
\textbf{ACC} & $0.7514$ & $0.7699$ & $0.7786$ & $0.7828$ & $0.9279$ \\
\hline
\end{tabular}
\caption{\label{tab:baseline} \textbf{Performance of state-of-the-art algorithms.}}
\end{table}

\begin{figure}[h!]
\centering
\subfigure[PRE@K]{
\includegraphics[width=0.23\textwidth]{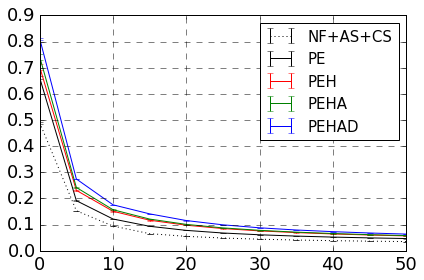}}
\hspace{-8pt}
\subfigure[REC@K]{
\includegraphics[width=0.23\textwidth]{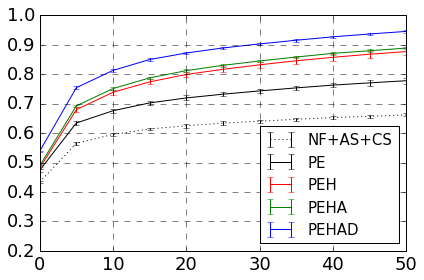}}
\vspace{-5pt}
\caption{\textbf{Performance of our metric learning methods.}}
\vspace{-5pt}
\label{fig:sup}
\end{figure}

\begin{table}[t!]
\small
\centering
\begin{tabular}{|c|c|c|c|}
\hline
\textbf{Method} & \textbf{avg.~PRE} & \textbf{avg.~REC} & \textbf{ACC}\\ 
\hline
\textbf{NF+AS+CS} & $ .05508\pm.00024 $ &  $ .6476\pm .0058$ & $ .6989\pm .0135$ \\
\hline
\textbf{PE} & $.07533 \pm .00034$ &  $.7534 \pm .0083$ & $. 7926\pm .0297$ \\
\hline
\textbf{PEH} & $ .08970\pm .00039 $ &  $ .8434\pm.0102 $ & $ .8580\pm.0238 $ \\
\hline
\textbf{PEHA} & $ .09274\pm.00019 $ &  $ .8554\pm.0047 $ & $ .8815\pm.0100 $ \\
\hline
\textbf{PEHAD} & $ .10780\pm .00021$ &  $ .9119\pm.0062 $ & $ .9279\pm .0104$ \\
\hline
\end{tabular}
\caption{\label{tab:sup} \textbf{Performance of our metric learning methods.}}
\end{table}

\begin{figure}[h!]
\centering
\vspace{-10pt}
\subfigure[Training loss]{
\includegraphics[width=0.23\textwidth]{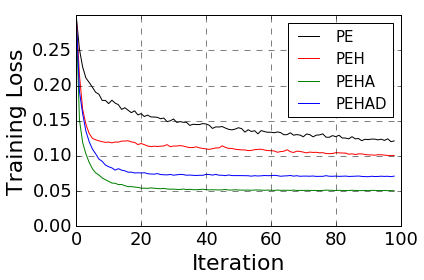}}
\hspace{-8pt}
\subfigure[Testing accuracy]{
\includegraphics[width=0.23\textwidth]{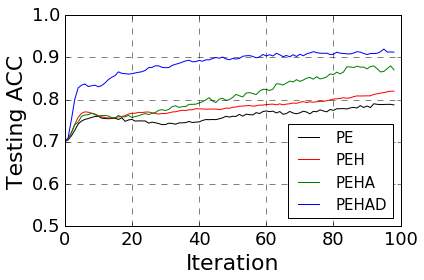}}
\vspace{-5pt}
\caption{\textbf{Training loss and testing accuracy curves.}}
\vspace{-5pt}
\label{fig:curve}
\end{figure}

%As for our models, we mildly tuned the hyper-parameters through standard five-fold cross-validation. For the basic models, we set the neural architecture of FNN to $400\to300\to200$, the batch size $|B|$ to $2^{16}$, and the learning rate $r$ to 0.001; For the basic models, we set the marge parameter $\alpha$ to 0.5; For hard sampling, we set the slack parameter $\beta$ to 0.2; For attentive training, we set the weighting parameter $\lambda$ of regularization loss to 1; For label denoising, we set the number of clusters to $K$ to $1,000$, and the weighting parameter $\rho$ of denoising loss to 0.1. All models are implemented with PyTorch and optimized through standard gradient backpropagation with mini-batch Adam \cite{kinga2015method}.

Figure \ref{fig:sup} and Table \ref{tab:sup} show the performance of compared models. As we can see, our supervised metric learning framework can significantly improve the place embedding quality,  because labeled data allow the models to pick out the more important features that differentiate duplication pairs from non-duplications. Moreover, each of our proposed model components further boosts the overall model performance, and their coherent combination (\textit{PEHAD}) has the best performance regarding all metrics in the evaluation. Our models also significantly outperform all of the state-of-the-art non-embedding algorithms in Table \ref{tab:baseline}-- \textit{PEHAD} achieves over 18\% relative improvement on \textit{GBDT} (the strongest baseline).

To better understand how each of the components contributes to the overall model, we also closely observe the training loss and testing accuracy during the training of different models. As we can see in Figure \ref{fig:curve}: 1) Hard sampling is effective in speeding up the convergence of training loss, and allows the model to keep learning after the loss seems to converge; 2) Attentive training leads to lower losses due to the additional weights. It does not lead to significant performance gain on top of the basic \textit{PE} model with hard sampling, but it does make the training process more stable and reduce the performance variance across different trains of the same model; 3) Label denoising empowers the model to rapidly remove the influence of noisy labels and achieve the peak performance after a small number of training iterations, which may further allow efficient model training with early stop.

Finally, all techniques here are also not bounded to the problem of place deduplication, but can rather be combined to leverage noisy pair-wise labels from multiple sources in various other domains.

%!TEX root = dedup.tex
\section{Conclusions}
\label{sec:con}
In this paper, we collect data from the real-world place graph of Facebook as an example to systematically study the novel problem of place deduplication and comprehensively evaluate the effectiveness of our proposed place embedding pipeline. 
%The concepts and methods developed are general and promising for any platform with similar types of data regarding object attributes and training labels.
The concepts and methods developed, although do not provide an end-solution for place deduplication, successfully improve the quality of place embedding for intermediate tasks like duplication prediction and candidate fetch, which largely facilitates further place deduplication. 

As for future work, it is interesting to further improve both the unsupervised feature generation model to incorporate more attributes, and the supervised metric learning model to leverage various training signals with efficiency. 
It is also interesting to apply the learned place embedding to facilitate other tasks related to place graph quality such as junk place prediction and common place clustering, as well as downstream applications including next destination recommendation and location-based ads ranking.

\bibliographystyle{ACM-Reference-Format}
\bibliography{carlyang} 
\end{document}